\documentclass{article}
\usepackage{cite}
\usepackage{ijcai16}
\usepackage{algorithm}
\usepackage{algorithmic}
\usepackage{graphicx}
\usepackage{amsmath}
\usepackage{caption} 
\usepackage{times}

\newcommand{\argmax}{\arg\!\max}

\title{Algorithms for Batch Hierarchical Reinforcement Learning}
\author{Tiancheng Zhao and Mohammad Gowayyed \\
 Language Technologies Institute, Carnegie Mellon University\\
\{tianchez, gowayyed\}@cs.cmu.edu}

\begin{document}

\maketitle

\begin{abstract}
Hierarchical Reinforcement Learning (HRL) exploits temporal abstraction to solve large Markov Decision Processes (MDP) and provide transferable subtask policies. In this paper, we introduce an off-policy HRL algorithm: Hierarchical Q-value Iteration (HQI). We show that it is possible to effectively learn recursive optimal policies for any valid hierarchical decomposition of the original MDP, given a fixed dataset collected from a flat stochastic behavioral policy. We first formally prove the convergence of the algorithm for tabular MDP. Then our experiments on the Taxi domain show that HQI converges faster than a flat Q-value Iteration and enjoys easy state abstraction. Also, we demonstrate that our algorithm is able to learn optimal policies for different hierarchical structures from the same fixed dataset, which enables model comparison without recollecting data.
\end{abstract}

\section{Introduction}

Conventional tabular reinforcement learning is bottle-necked by the curse of dimensionality for practical applications. The number of parameters that needs to be trained grows exponentially with respect to the size of states and actions. In order to make reinforcement learning practically tractable, one line of research is hierarchical reinforcement learning (HRL), which develops principled ways of temporal and state abstraction to reduce the dimensionality for sequential decision making.

The basic idea of temporal abstraction is to develop macro-actions that take several steps to terminate before returning. Usually good macro-actions aim to solve sub-goals, so that multiple macro-actions divide difficult tasks into several simpler ones. In addition, state abstraction tries to reduce the dimensionality by removing irrelevant state variables for decision making, reducing the cardinality of state space and helping in tackling over-fitting. These two techniques lead to natural hierarchical control architecture, which intuitively resembles how humans solve complex tasks.

Another area of research closely related to our work is batch reinforcement learning. Batch reinforcement learning aims to learn the best policy from a fixed set of prior-known samples. Compared to on-policy algorithms, batch reinforcement learning enjoys stability and data-efficiency. More importantly, it allows to apply reinforcement learning in a practical problem that is expensive in collecting new samples, such as education, spoken dialog system and medical systems. Well-known algorithms in batch reinforcement learning include Least Square Policy Iteration (LSPI) \cite{lagoudakis2003least}, Fitted Q iteration (FQI) \cite{ernst2005tree}, Neural Fitted Q Iteration (NFQ) \cite{riedmiller2005neural} and etc. 

In this paper, we combine batch learning and hierarchical reinforcement learning, in order to achieve faster learning speed, data efficiency and model comparison.

\section{Related Work}

There are three major approaches developed relatively independently \cite{barto2003recent}, aiming to formalize the idea of abstraction into reinforcement learning. The three approaches are: 1) the \textit{option} framework \cite{sutton1999between}, 2) \textit{Hierarchies of Abstract Machines} (HAMs) \cite{parr1998reinforcement} and 3) MAXQ framework \cite{dietterich2000overview}.

Under the \textit{option} framework, the developers augment the original action set by \textit{options}, which are macro actions that have their own predefined policy, termination state and active state. Sutton et al have shown that such a system is a semi-Markov Decision Process (SMDP), which converges to a unique hierarchical optimal solution using a modified Q-learning algorithm. 
For HAM framework, rather than giving out the entire policy of these macro actions, developers just need to specify a \textit{partial program} that specifies a part of the policy. Using HAMQ learning \cite{parr1998reinforcement}, HAM can also converge to a hierarchical optimal solution.

At last, MAXQ framework provides an elegant formulation that decomposes the original MDP into several subroutines in a hierarchy and the algorithm can learn policies recursively for all the subroutines. Therefore, in the MAXQ framework, there is no need to specify the policy for any macro-actions. However, Dietterich shows that it can only achieve recursive optimal solution, which in the extreme case, can be arbitrarily worse than the hierarchical optimal solution. 

All of the above work assumes that the agent can interact with the world while learning. However, in real-world applications that needs HRL, it is usually very expensive to collect data and terrible failures are not allowed on operation. This forbids the usage of online learning algorithms that could potentially preform horribly in the early learning stage. To our best knowledge, there is little prior work~\cite{bloch2011reducing} in developing batch learning algorithms that allow a hierarchical SMDP to be trained from an existing data set collected from a stochastic behavior policy. We believe that such algorithms are valuable for applying HRL in complex practical domains. 

\section{Batch Learning for HSMDP}

\subsection{Definitions}
Mostly, we follow the definitions in the MAXQ framework. However, for notation simplicity, we also borrow some notations from the $option$ framework. 

\subsection{Markov Decision Process}
An MDP $M$ is described by $(S, A, P, R, P_0)$ 
\begin{enumerate}
\item $S$ is the state space of $M$
\item $A$ is a set of primitive actions that are available 
\item $P(s'|s,a)$ defines the transition probability of executing primitive action $a$ in state, $s$
\item $R(s'|s,a)$ is the reward function defined over $S$ and $A$ 
\end{enumerate}

\subsection{Hierarchical Decomposition}
An MDP, $M$, can be decomposed into a finite set of subtasks $O=\{O_0, O_1...O_n\}$ with the convention that $O_0$ is the root subtask, i.e. solving $O_0$ solves the entire original MDP, $M$. $O_i$ is then a Semi-Markov Decision Process (SMDP) that shares the same $S$, $R$, $P$ with $M$, and has an extra tuple $<\beta_i, U_i>$, where:
\begin{enumerate}
\item $\beta_i(s)$ is the termination predicate of subtask $O_i$ that partition $S$ into a set of active states, $S_i$ and a set of terminal states $T_i$. If $O_i$ enters a state in $T_i$, $O_i$ and its subtasks exit immediately, i.e. $\beta_i(s)=1$ if $s\in T_i$, otherwise $\beta_i(s)=0$.
\item $U_i$ is a nonempty set of actions that can be performed by $O_i$. The actions can be either primitive actions from $A$ or other subtask, $O_j$, where $i\neq j$. We will refer to $U_i$ as the children of subtask $O_i$.
\end{enumerate}
It is evident that a valid hierarchical decomposition forms a direct acyclic graph (DAG) where each non-terminal node corresponds to a subtask, and each terminal node corresponds to a primitive action. For later discussion, we will use \textit{hierarchical decomposition} and \textit{DAG} interchangeably.

\subsection{Hierarchical Policy}
A hierarchical policy, $\pi$, is a set of policies for each subtask, $O_i$, $\pi=\{\pi_0, \pi_1...\pi_n\}$. In the terminology of $option$ framework, a subtask policy is a deterministic $option$, with $\beta_i(s)=1$ for $s\in T_i$, and $0$ otherwise.

\subsection{Recursive Optimality}
A \textit{recursive optimal} policy for MDP $M$ with hierarchical decomposition is a hierarchical policy $\pi=\{\pi_0...\pi_n\}$, such that for each subtask, $O_i$, the corresponding policy $\pi_i$ is optimal for the SMDP defined by the set of states, $S_i$, the set of actions $U_i$, the state transition probability $P^{\pi}(s', N|s,a)$, and the rewards function $R(s'|s,a)$.

\section{Algorithm}
The problem formulation is as following: given any finite set of samples, $D=\big\{ (s_m, a_m, r_m, s^{'}_m) |m=1,2,...,M \big\}$ and any valid hierarchical decomposition $O$ of the original MDP $M$, we wish to learn the recursive optimal hierarchical policy $\pi^{*}$.

We now propose Hierarchical Q-value Iteration, HQI, and we prove that it converges to the recursive optimal solution for any hierarchical decomposition given that the batch sample distribution has sufficient state action exploration. The basic idea is to train every subtask using Subtask Q-value Iteration (SQI) in a bottom up fashion. The training prerequisite of SQI for a specific subtask $O_i$ is that all of its children $U_i$ have converged to their greedy optimal policies. In order to fulfil this constraint, HQI first topologically sorts the \textit{DAG} and running SQI from subtasks whose children have only primitive actions. After those subtasks converge to their optimal policy, the algorithm continues to other subtasks whose children are either converged or primitive actions. We will show that there always exist an ordering of training every subtask in a valid \textit{DAG} that fulfills the prerequisite of SQI.

One challenge of training a subtask with subtask children is that we cannot use the optimal SMDP Bellman equation described in the MAXQ framework~\cite{dietterich2000overview}, $Q_i(s, u)$, which is the Q-value function for subtask, $O_i$, at state, $s$ and action $u$: 
\begin{equation}
\label{eq:maxqq}
Q_i(s, u) = V(s, u) + \sum_{s', N} P_i^{\pi}(s', N | s, u)\gamma^N Q_i^{\pi}(s', \pi_i(s')) \\
\end{equation}
\begin{equation}
V(s, u) = \begin{cases}
        max_{u'}(Q_u(s, u')) &\text{$u$ is subtask}\\
        \sum_{s'} P(s'|s,u)R(s'|s,u) &\text{$u$ is primitive}
        \end{cases}
\end{equation}
The main problem of this equation\eqref{eq:maxqq} is that in order to estimate the Q-value for a subtask children, $u$, the parent $O_i$ needs to have an estimate about the transition probability $P_i^{\pi}(s', N | s, u)$, which is the distribution of $u$'s exit state and number of primitive steps needed to reach its termination. Although the termination states of the child $u$ are given by $T_u$, it is difficult to estimate the joint distribution of termination steps $N$ and $s$' if $u$ follows an policy that is different from the behavior policy without recollecting new samples. This is because since the behavior policy is usually random and poor in performance, the collected samples do not provide information about how many steps the subtask $u$ would take to terminate if following a different (optimal) policy.

Therefore, instead of using the above Bellman equation that updates the Q table of the parent when a child exits, we use the intra-option Bellman equation proposed in the $option$ framework~\cite{sutton1999between}:
\begin{align}
\label{eq:optionq}
Q_i(s, u) &= \sum_{a\in A}{\pi_u(a|s)E[r(s,a) + \gamma V_i(s',u)]}\\
 &= \sum_{a\in A}{\pi_u(a|s)\bigg [r(s,a) + \gamma \sum_{s'}P(s'|s,a)V_i(s',u)\bigg]}
\end{align}
Where 
\begin{equation}
V_i(s,u) = (1-\beta_i(s))Q_i(s, u) + \beta_i(s)\max_{u'\in U_i} Q_i(s, u')
\end{equation}

Equation \eqref{eq:optionq} also yeilds a contraction in the max norm and is able to learn the Q table after observing every new reward, which eliminates the need to estimate $P_i^{\pi}(s', N | s, u)$. Another key benefit is that we can use flat samples to estimate the one step transition probability and rewards in equation \eqref{eq:optionq}, which makes the algorithm independent of the hierarchical decomposition and is able to learn optimal polices for different structures from the same dataset. Specifically, we can estimate the above two terms by $\sum_{s'}P(s'|s,a)V(i, s',u) \approx \frac{1}{c}\sum_{m=1}^{c}V(i, s'_{m}=s', u)$ and $r(s,a) \approx \frac{1}{c}\sum_{m=1}^{c}r(s_m=s, a_m=a)$, where $c$ is the number of experiences that has $s'$ and ($s$, $a$), respectively. At last, since we assume converged subtasks follow deterministic greedy policy, $\pi_u(a|s) = 1$ if $a$ is the greedy primitive action that subtask $u$ would take at state $s$, and $\pi_u(a|s) = 0$ otherwise. This step is in fact crucial for HQI to learn the optimal policy because it allows a subtask to discard those samples that are not following the optimal behavior of its children.

The HQI algorithm is summarized in Algorithm~\ref{alg:hqi} and SQI is summarized in Algorithm~\ref{alg:sqi}. Any dataset $D$ can be used at every iteration of SQI. If the initial data is sufficient to cover important state-action space, the same dataset is able to train all subtasks of the \textit{DAG}. 

\subsection{Extension to Function Approximation and State Abstraction}
We note that it is trivial to extend SQI to Fitted-SQI which uses a function approximator to model the Q-value function for a subtask at the end of each iteration. The direct advantage of using function approximation is that it can incorporate powerful supervised regression methods, such as Gaussian Processes or Neural Networks to scale up to large-scale and continuous MDPs. Although using function approximations usually compromises the theoretical convergence guarantee for tabular MDP, our experiments shows that Fitted-HQI is able to converge to the unique optimal solution. The Fitted-SQI is summarized in Algorithm~\ref{alg:fsqi}.

Furthermore, state abstraction here means finding a subset of state variables that are most informative for each subtask. A good hierarchical decomposition decomposes the original MDP into several simpler ones, such that the agent only needs to care about a small set of features in each task. Therefore, a good structure should create easy opportunity for state abstraction at each level of its hierarchy. Many techniques have been explored~\cite{painter2012greedy}\cite{qin2014sparse}\cite{geramifard2013batch} in non-hierarchical batch reinforcement learning to achieve state abstractions. These methods can be directly applied in fitting step of Fitted-SQI and each subtask can learn its own sparse state representation. Due to the space limit, we conduct simple manual state abstraction for each subtask in this paper, and leave the study of analyzing automatic feature selection techniques into future works.

\begin{algorithm}
\caption{Hierarchical Q-value iteration (HQI)}
\begin{algorithmic} 
\label{alg:hqi}
\REQUIRE $O$, $D$
\STATE $Train \leftarrow O_i \in O$ with only primitive children
\STATE $Done \leftarrow \{A\}$
\WHILE{$Train\neq empty$}
    \FOR{$O_i\in Train$} 
        \STATE{SQI$(O_i, D)$} 
        \STATE{$Done.add(O_i)$}
    \ENDFOR
    \STATE $Train \leftarrow O_i \in (O-Done)$ AND $U_i \in Done$  
\ENDWHILE
\end{algorithmic}
\end{algorithm}

\begin{algorithm}
\caption{Subtask Q-value Iteration (SQI)}
\begin{algorithmic} 
\label{alg:sqi}
\REQUIRE $O_i,D$
\WHILE{$k < maxIter$}
    \FOR{$(s,a,r,s') \in D$}
        \FOR{$u \in U_i$}
            \IF{$s' \in T_i$}
                \STATE $y \leftarrow r$
                \STATE $Q^k_{i}(s, u) \leftarrow (1-\alpha)Q_i^{k-1}(s, u) + \alpha y$
            \ELSE
                \IF{GreedyPolicy($u$, $s$) $=a$}
                    \STATE $y \leftarrow r + \gamma((1-\beta_u(s')) Q_i^{k-1}(s', u)$
                    \STATE $\quad \quad \quad + \beta_u(s')max_{u'\in U_i}Q_i^{k-1}(s', u'))$
                    \STATE $Q_i^{k}(s, u) \leftarrow (1-\alpha)Q_i^{k-1}(s, u) + \alpha y$
                \ENDIF
           \ENDIF
        \ENDFOR
    \ENDFOR
\ENDWHILE
\end{algorithmic}
\end{algorithm}

\begin{algorithm}
\caption{Fitted Subtask Q-value Iteration (Fitted SQI)}
\begin{algorithmic} 
\label{alg:fsqi}
\REQUIRE $O_i,D$
\WHILE{$k < maxIter$}
    \STATE{$X\leftarrow []$, $Y\leftarrow []$}
    \FOR{$(s,a,r,s') \in D$}
        \FOR{$u \in U_i$}
            \IF{$s' \in T_i$}
                \STATE $y \leftarrow r$
            \ELSE
                \IF{GreedyPolicy($u$, $s$) $=a$}
                    \STATE $y \leftarrow r + \gamma((1-\beta_u(s')) Q_i^{k-1}(s', u)$
                    \STATE $\quad \quad \quad + \beta_u(s')max_{u'\in U_i}Q_i^{k-1}(s', u'))$
                \ENDIF
           \ENDIF
           \STATE{$X.add((s, u))$, $Y.add(y)$}
        \ENDFOR
    \ENDFOR
    \STATE{$Q_i^{k} \leftarrow fit(X, Y)$}
\ENDWHILE
\end{algorithmic}
\end{algorithm}

\begin{algorithm}
\caption{GreedyPolicy}
\begin{algorithmic} 
\REQUIRE $u, s$
\IF{$u \in A$}
    \RETURN u
\ELSE
    \STATE $u^*=\argmax_{u' \in U_u} Q_u(s, u')$
    \RETURN GreedyPolicy($u^*$, $s$)
\ENDIF
\end{algorithmic}
\end{algorithm}

\section{Proof of Convergence}
In this section, we prove that our HQI (in the tabular case) converges to the recursive optimal policy. Assume that the policy at each subtask $M_i$ is ordered, such that it break ties deterministically (e.g favor left to right), it defines a unique recursive optimal hierarchical policy, $\pi^*_r$, and a corresponding recursive optimal Q function $Q^*_r$. We then show that HQI converge to  $\pi^*_r$ and $Q^*_r$. The $r$ subscript refers to recursive optimality. 

\subsection{Proof}
We want to prove that for an MDP $M=(S,A,P,R,P_0, \gamma)$ with hierarchical decomposition $O=\{O_0,..O_n\}$, HQI converges to recursive optimal policy for the hierarchical policy of $M$, $\pi^*_r$.

\begin{enumerate}
\item We first prove that: For a subtask $O_i$, with all of its children converged to their recursive optimal policies and infinity amount of batch data, algorithm SQI converge to the optimal Q-value function after infinity number of iterations,  $\lim_{k->\infty}Q^k_i=Q^*_i$

\item We then show that HQI provides an order of training all the subtasks in the \textit{DAG} graph, such that when training a subtask, $O_i$, all of its children, $U_i$ already converged to their optimal recursive policies.

\end{enumerate}

\subsection{Proof Sketch}
\textbf{Step 1}
We begin with the base case for subtask whose children are all primitive actions. We can notice that equation~\eqref{eq:optionq} falls back to traditional Bellman operator for flat MDP, because a primitive action always terminate after one step:
\begin{equation}
Q(i, s, u) = r(s,a) + \gamma \sum_{s'}P(s'|s,a)\max_{u'\in U_i} Q(i, s, u')
\end{equation}
Therefore, for subtask with only primitive children, SQI is equivalent to flat Q-value iteration, which is guaranteed to converge to optimal policy given sufficient data.

Then for subtasks with other subtask children, by definition, when we run SQI, the children of $O_i$ ($U_i$)  have converged to their unique deterministic optimal recursive policy. This means that every action $u \in U_i$, is a deterministic \textit{deterministic  Markov option} as defined in the $option$ framework \cite{sutton1999between}. \cite{sutton1999between} proved that "for any set of \textit{deterministic Markov options} one step intra-option Q-learning converges w.p. 1 to the optimal Q-values, for every option regardless of what options are executed during learning provided every primitive action gets executed in every state infinitely often". Refer to the \cite{sutton1999between}, for the detailed proof.

\textbf{Step 2} By definition, a hierarchical decomposition is a Directed Acyclic Graph (DAG) with edges from parents to their children. In this proof, we first reverse the edges so that they are from children to their parents. Also we know from Graph Theory that any Directed Acyclic Graph has at least one topological sort, such that every edge $uv$, $u$ comes before $v$ in the ordering~\cite{cormen2001section}. Therefore, we can to topologically sort the hierarchical decomposition with reversed edges such that SQI can always train the children before parents.

Also, the definition of topological sort ensures the initial condition that there is at least one subtask that only has primitive children. Therefore, we can then conclude that for any DAG, HQI can traverse the subtasks such that the conditions of SQI convergence are met. Then, HQI converges for all subtasks.

\section{Experiments}

\subsection{Experimental Setup}

\begin{figure}
\begin{centering}
    \includegraphics[width=0.25\textwidth]{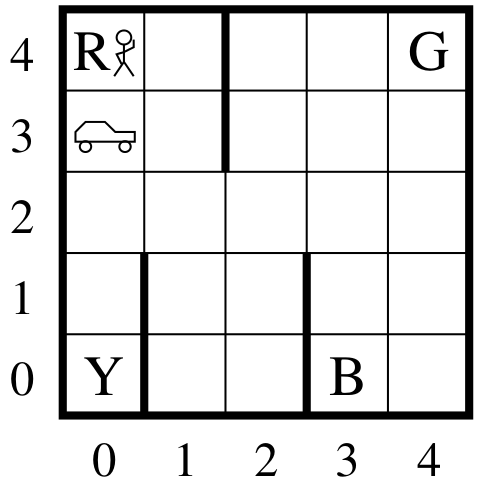}
    \caption{The Taxi Domain}
    \label{fig:taxi0}
\end{centering}
\end{figure}

We applied our algorithm to the Taxi domain described in \cite{sutton1999between}. This is a simple grid world that contains a taxi, a passenger, and four specially-designated locations labeled R, G, B, and Y. In the starting state, the taxi is in a randomly-chosen cell of the grid, and the passenger is at one of the four special locations. The passenger has a desired destination that he/she wishes to reach, and the job of the taxi is to go to the passenger, pick him/her up, go to the passenger's destination, and drop the passenger. The taxi has six primitive actions available to it: move one step to one of the four directions (north, south, east and west), pick up the passenger and put down the passenger. To make the task more difficult, the move actions are not deterministic, so that it has $20\%$ chance of moving in one of the other directions. Also, every  move in the grid will cost $-1$ reward. Attempting to pick up or drop passenger at wrong location will cause $-10$ reward. At last, successfully finish the task has $20$ reward. The grid is described in figure \ref{fig:taxi0}. Therefore, there are 4 possible state for the destination, 5 possible state for the passenger (4 location and 5 is on the car), 25 possible locations, which results into $500*6=3000$ parameters in the Q-table that needs to be learned. We denote the state variable as $[dest, pass, x, y]$ for later discussion.

The dataset for each run were collected in advance by choosing actions uniformly at random with different sizes. We evaluate the performance of algorithms by running greedy execution for $100$ times to obtain average discounted return at every $5000$ new samples and up to $60,000$ samples. We repeat the experiments for $5$ times to evaluate the influence of different sample distribution. The discounting factor is set to be $0.99$.

\begin{figure}
\begin{centering}
    \includegraphics[width=0.46\textwidth]{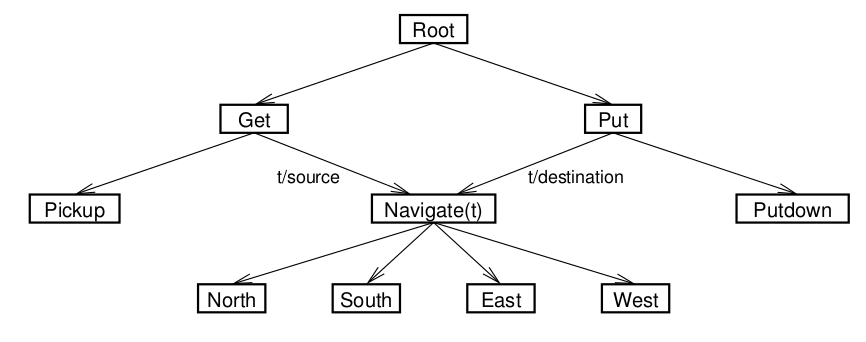}
    \caption{DAG 1}
    \label{fig:DAG1}
\end{centering}
\end{figure}

\subsection{Results}
We conducted three sets of experiments: 1) comparison of HQI with flat Q-value Iteration and the effect of state abstraction. 2) learning polices for different DAGs from the same dataset and 3) learning policy using Fitted-HQI with Random Forest as the function approximator.

\begin{figure}
    \includegraphics[width=0.46\textwidth]{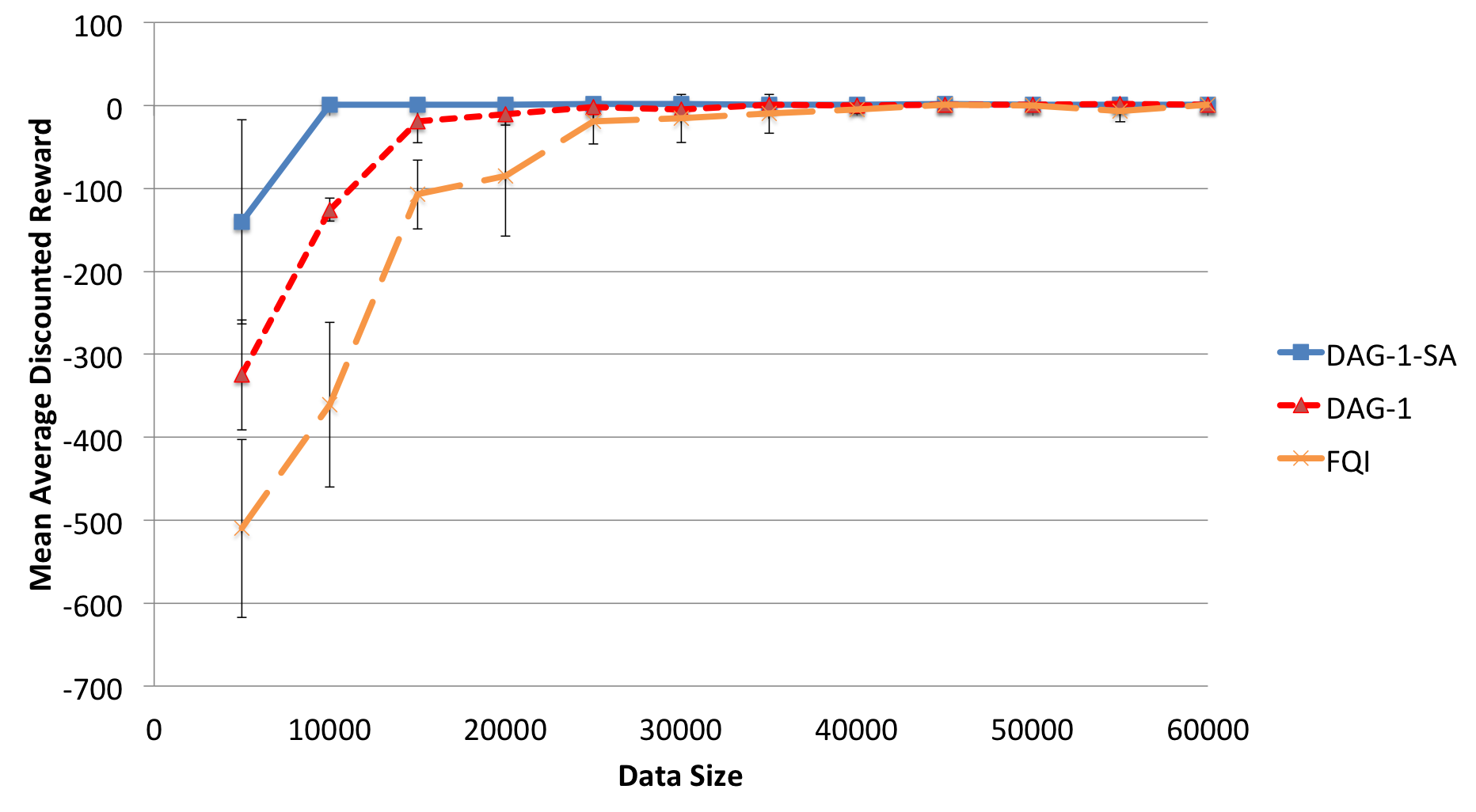}
  \caption{Average discounted reward. The SA postfix means state abstraction. The error bar is one standard deviation over 5 runs.}
  \label{fig:res0}
\end{figure}

\begin{figure}
    \includegraphics[width=0.46\textwidth]{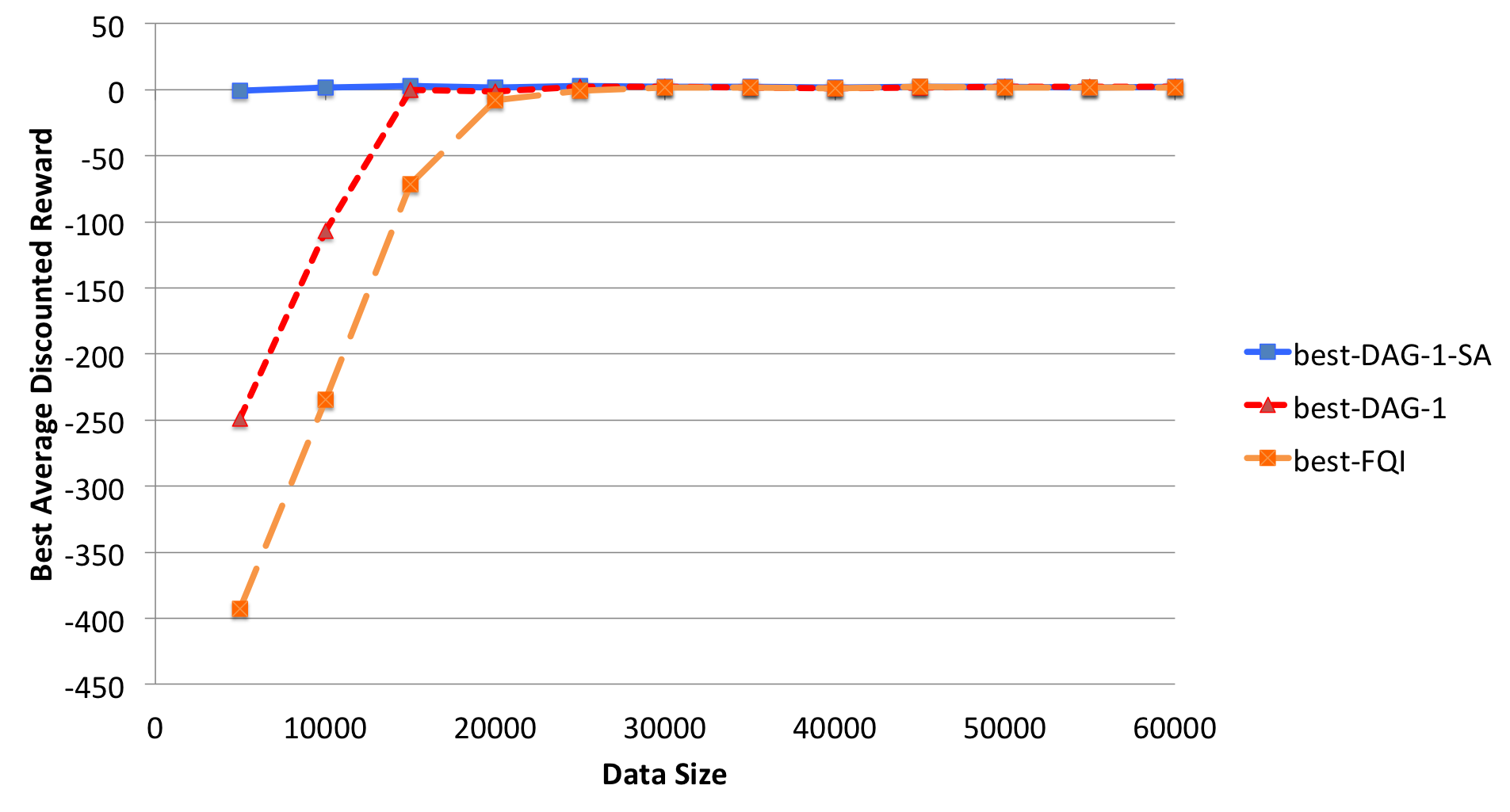}
  \caption{Best performance comparison}
  \label{fig:res1}
\end{figure}

The \textbf{first} experiment compares HQI against flat Q-value Iteration (FQI). Also, as pointed out in~\cite{dietterich2000hierarchical}, state abstraction is essential for MAXQ to have fast learning speed compared to flat Q learning. As a result, we manually conduct state abstraction for each subtask in DAG $1$. However, different from the aggressive state abstraction described in~\cite{dietterich2000hierarchical}, where every subtask and child pair has a different set of state variables, we only conduct a simple state abstraction at subtask level, i.e. all children of a subtask has the same state abstraction. The final state abstraction is listed in Table~\ref{tbl:DAG1}. As described above, we run $5$ independent runs with different random samples of different sizes, we report the mean average discounted return over five runs in Figure~\ref{fig:res0}, as well as the best average discounted reward of the five runs in Figure~\ref{fig:res1}. 

\begin{table}[]
\centering
\caption{DAG-1 State Abstraction}
\label{tbl:DAG1}
\begin{tabular}{|l|l|}
\hline
\textbf{subtask}   & \textbf{active states}  \\ \hline
root      & {[}pass{]}     \\ 
get       & {[}pass x y{]} \\
put       & {[}dest x y{]} \\ 
navi\_get & {[}pass x y{]} \\
navi\_put & {[}dest x y{]} \\ \hline
\end{tabular}
\end{table}

Results show that both HQI with and without state abstraction consistently outperforms the FQI when there is limited training data. When the dataset is large enough, they all converge to the same optimal performance, which is around $1.0$. We also notice that, occasionally, HQI with state abstraction can learn the optimal performance state abstraction with very limited samples, i.e $5000$ samples. This demonstrates that with proper hierarchy constraints and good behavioral policy, HQI can generalize much faster than FQI. Moreover, even the HQI without state abstraction consistently outperforms FQI in terms of sample efficiency. This is different from the behavior of the on-policy MAXQ-Q algorithm reported in~\cite{dietterich2000hierarchical}, which needs state abstraction in order to learn faster than Q-learning. We argue that HQI without state abstraction is more sample efficient than FQI for the following reasons: 1) HQI uses all applicable primitive samples to update the Q-table for every subtask while MAXQ-Q only updates for the subtask that executes that particular action. 2) Upper level subtask in MAXQ-Q needs to wait for its children gradually converges to their greedy optimal policy before it can have have a good estimate of $P(s', N|s, u)$ while HQI does not have this limitation.

\begin{figure}
\begin{centering}
    \includegraphics[width=0.35\textwidth]{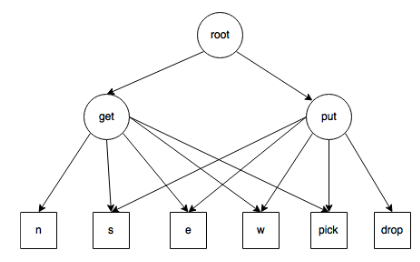}
    \caption{DAG 2}
    \label{fig:DAG2}
\end{centering}
\end{figure}

\begin{figure}
\begin{centering}
    \includegraphics[width=0.35\textwidth]{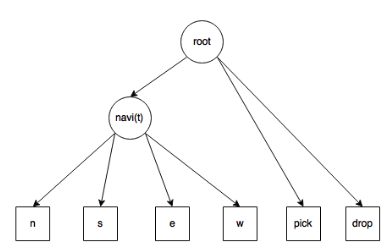}
    \caption{DAG 3}
    \label{fig:DAG3}
\end{centering}
\end{figure}

\begin{figure}
    \includegraphics[width=0.46\textwidth]{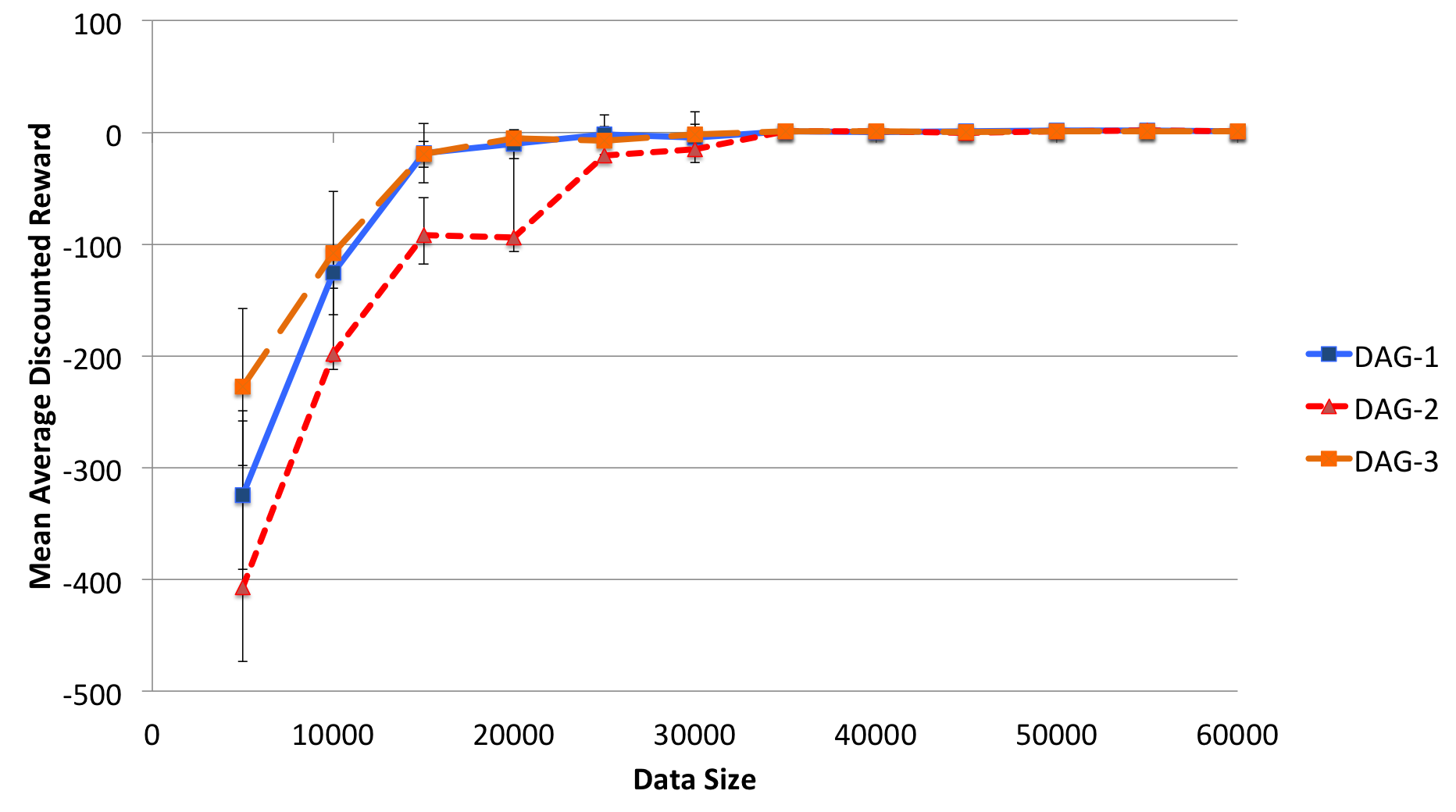}
  \caption{Comparison of different DAGs. The error bar is one standard deviation over 5 runs.}
  \label{fig:res2}
\end{figure}

The \textbf{second} experiment is running HQI on different variations of hierarchical decomposition of the original MDP. Figure~\ref{fig:DAG2} and Figure~\ref{fig:DAG3} show two different valid DAGs that could also solve the original MDP. Figure~\ref{fig:res2} demonstrates that with sufficient data all three DAG converge to their recursive optimal solution, which confirms that HQI is able converge for different hierarchies. In terms of sample efficiency, three structures demonstrate slight different behavior. We can notice that DAG $2$ learns particularly slower than the other two. We argue that this is because of poor decomposition of the original MDP. Based on the problem settings, \textit{pick} and \textit{drop} are all risky actions (illegal execution lead to $-10$ reward), while in DAG $2$ these two actions are mixed with low-cost \textit{move} actions while the other two DAGs isolated them in a higher level of decision making. Therefore, designing good hierarchy is crucial to obtain performance gain versus flat RL approaches. This  emphasizes the importance of the off-policy nature of HQI, which allows developers to experiment with different DAG structures without collecting new samples. How to effectively evaluate the performance of particular hierarchical decomposition without using a simulator is a part of our future research.

\begin{figure}
    \includegraphics[width=0.46\textwidth]{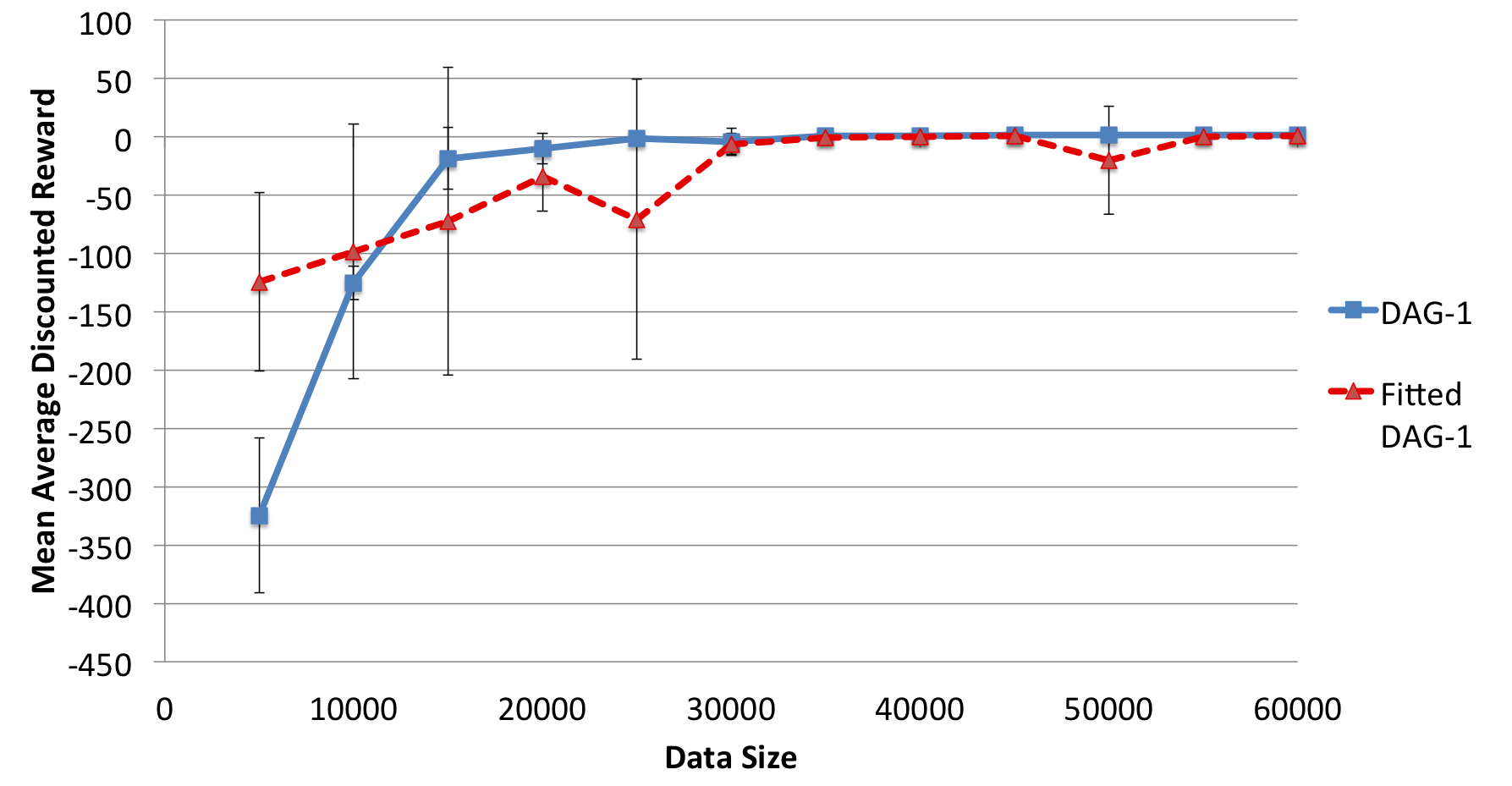}
  \caption{Comparison between fitted-HQ and HQI on DAG-1. The error bar is one standard deviation over 5 runs.}
  \label{fig:res3}
\end{figure}

The \textbf{last} experiment utilizes Random Forests as the function approximator to model the Q-value function in DAG 1. The main purpose is to demonstrate the convergence of Fitted HQI. For each subtask $O_i$ the Q-value function $Q_i(s, u)$ is modelled by a random forest with $[dest, pass, x, y]$ as the input feature. Since $dest$ and $pass$ are categorical variables, we represent them as a one-hot vector, which transforms the state variable into a $11$ dimension vector (4d for destination, 5d for passenger and 2d for the $x,y$ coordinate). We report the mean average discounted rewards over 5 independent runs with different random samples of different sizes. Figure~\ref{fig:res3} shows that Fitted-HQI achieves similar performance compared to Tabular HQI.

\subsection{Comparison with MAXQ-Q and Intra-Option Learning}
Compared to MAXQ-Q, HQI enjoys sample efficiency and the ability to be off-policy. The advantage of off-policy is that it does not require hyper-parameter tuning such as exploration rate. Since high level subtask in MAXQ-Q needs to wait for its children converge first, developers usually set a faster exploration decay rate for lower level subtasks, which is an extra hyperparameter that needs tuning. The limitation of HQI is that it maintains an independent Q table for each subtask, while MAXQ-Q allows a part of the parent value function recursively retrieved from its children, a technique known as value function decomposition~\cite{dietterich2000hierarchical}. This allows more compact memory usage and accelerates the learning of the parents. How to share value function in the off-policy setting is a future research topic.

For intra-option learning in \textit{option} framework, the main advantage of HQI is that it does not require developers to fully define the policy of every options. Instead, one only needs to define a DAG with terminal predicate for each node in the graph. We argue that in general it is easier to define a task hierarchy than giving a full policy for macro-actions. Therefore, HQI combines the strength of intra-option off-policy learning with MAXQ. \cite{bloch2011reducing} provides a method of training options in an off-policy fashion. Compared to it, HQI has the advantage of learning all subtasks from flat batch dataset, so that our algorithm does not require a task DAG priory to collecting data and a manual definition of option policies.

\section{Conclusion and Future Work}
In this paper, we introduced an off-policy batch learning algorithm for hierarchical RL. We showed that it is possible to blindly collect data using a random flat policy. Then, we use this data to learn different structures that the data collection were not aware of. Our experiments on the Taxi domain show that it converges faster than FQI to the optimal policy. It also shows that different DAG structures are able to learn from this flat data, with different speeds. Every DAG structure has its own number of parameters, which suggests a possible line for research to try to minimize the number of parameters in the hierarchy. Other future work include comparing different feature selections techniques for Fitted-SQI and applying the algorithm to large-scale and complex domains.

\newpage

\bibliography{ijcai}{}

\begin{thebibliography}{10}

\bibitem{barto2003recent}
Andrew~G Barto and Sridhar Mahadevan.
\newblock Recent advances in hierarchical reinforcement learning.
\newblock {\em Discrete Event Dynamic Systems}, 13(1-2):41--77, 2003.

\bibitem{bloch2011reducing}
Mitchell~Keith Bloch.
\newblock Reducing commitment to tasks with off-policy hierarchical
  reinforcement learning.
\newblock {\em arXiv preprint arXiv:1104.5059}, 2011.

\bibitem{cormen2001section}
Thomas~H Cormen, Charles~E Leiserson, Ronald~L Rivest, and Clifford Stein.
\newblock Section 22.4: Topological sort.
\newblock {\em Introduction to Algorithms (2nd ed.), MIT Press and
  McGraw-Hill}, pages 549--552, 2001.

\bibitem{dietterich2000hierarchical}
Thomas~G Dietterich.
\newblock Hierarchical reinforcement learning with the maxq value function
  decomposition.
\newblock {\em J. Artif. Intell. Res.(JAIR)}, 13:227--303, 2000.

\bibitem{dietterich2000overview}
Thomas~G Dietterich.
\newblock An overview of maxq hierarchical reinforcement learning.
\newblock In {\em Abstraction, Reformulation, and Approximation}, pages 26--44.
  Springer, 2000.

\bibitem{ernst2005tree}
Damien Ernst, Pierre Geurts, and Louis Wehenkel.
\newblock Tree-based batch mode reinforcement learning.
\newblock In {\em Journal of Machine Learning Research}, pages 503--556, 2005.

\bibitem{geramifard2013batch}
Alborz Geramifard, Thomas~J Walsh, Nicholas Roy, and Jonathan How.
\newblock Batch-ifdd for representation expansion in large mdps.
\newblock {\em arXiv preprint arXiv:1309.6831}, 2013.

\bibitem{lagoudakis2003least}
Michail~G Lagoudakis and Ronald Parr.
\newblock Least-squares policy iteration.
\newblock {\em The Journal of Machine Learning Research}, 4:1107--1149, 2003.

\bibitem{painter2012greedy}
Christopher Painter-Wakefield and Ronald Parr.
\newblock Greedy algorithms for sparse reinforcement learning.
\newblock {\em arXiv preprint arXiv:1206.6485}, 2012.

\bibitem{parr1998reinforcement}
Ronald Parr and Stuart Russell.
\newblock Reinforcement learning with hierarchies of machines.
\newblock {\em Advances in neural information processing systems}, pages
  1043--1049, 1998.

\bibitem{qin2014sparse}
Zhiwei Qin, Weichang Li, and Firdaus Janoos.
\newblock Sparse reinforcement learning via convex optimization.
\newblock In {\em Proceedings of the 31st International Conference on Machine
  Learning (ICML-14)}, pages 424--432, 2014.

\bibitem{riedmiller2005neural}
Martin Riedmiller.
\newblock Neural fitted q iteration--first experiences with a data efficient
  neural reinforcement learning method.
\newblock In {\em Machine Learning: ECML 2005}, pages 317--328. Springer, 2005.

\bibitem{sutton1999between}
Richard~S Sutton, Doina Precup, and Satinder Singh.
\newblock Between mdps and semi-mdps: A framework for temporal abstraction in
  reinforcement learning.
\newblock {\em Artificial intelligence}, 112(1):181--211, 1999.

\end{thebibliography}
\bibliographystyle{plain}

\end{document}